%% file: root.tex
\documentclass[letterpaper, 10 pt, conference]{ieeeconf}  % Comment this line out if you need a4paper

\IEEEoverridecommandlockouts                              % This command is only needed if 
\usepackage{tabularx}
\usepackage{booktabs}
\usepackage{multirow}
\usepackage{graphicx}
\usepackage{cuted}
\usepackage{comment}
\usepackage{graphicx}
\usepackage{pgfplots}
\usepackage{xcolor} 
\usepackage[accsupp]{axessibility}
\let\labelindent\relax
\usepackage{enumitem}
\usepackage{booktabs}
\usepackage{array,multirow}
\usepackage{xspace}
\usepackage{booktabs} % for better-looking tables
\usepackage{graphicx} % for \scalebox
\usepackage{soul}
\usepackage{xcolor}
\usepackage[theorems,skins,breakable]{tcolorbox}
\usepackage{float}
\usepackage{silence}
\usepackage{pgfplots}
\usepackage{tikz}
\usepackage{xcolor}
\usepackage{amsmath} % or mathtools
\usepackage{caption}
\usepackage{graphicx}
\usepackage{subcaption}
\usepackage{float} % optional but helps with figure placement
\usepackage{hyperref}  % in preamble
\usepackage{amssymb}   % for \checkmark
\usepackage{pifont}    % for \ding
\usepackage{algpseudocode}
\usepackage{url}
\pgfplotsset{compat=1.18} % Use the latest compatibility for best results
\definecolor{mypurple}{RGB}{128,0,128} 
\definecolor{mygreen}{RGB}{0,128,0} 
\definecolor{mydarkblue}{RGB}{6,57,112} 
\definecolor{mylightblue}{RGB}{3, 165, 252}
\definecolor{hlightyellow}{RGB}{255,255,204}
\definecolor{lightyellow}{RGB}{255,255,204}
\definecolor{mypurple}{RGB}{100, 42, 150}
\definecolor{mypink}{RGB}{144, 71, 145} 
\definecolor{myred}{RGB}{143, 4, 7} 
\definecolor{mygreen}{RGB}{0,128,0} 
\definecolor{mygreen}{RGB}{0,128,0} 
\definecolor{white}{RGB}{255,255,255}

\title{\LARGE \bf
A Multi-Modal Neuro-Symbolic Approach for Spatial Reasoning-Based Visual Grounding in Robotics
}
\author{
    Simindokht Jahangard$^{1,*}$, 
    Mehrzad Mohammadi$^{2,*}$, 
    Abhinav Dhall$^{1}$, 
    and Hamid Rezatofighi$^{1}$%
    \thanks{*Equal contribution.}%
    \thanks{$^{1}$Faculty of Information Technology, Monash University, Australia. 
    {\tt\small \{simindokht.jahangard, abhinav.dhall, hamid.rezatofighi\}@monash.edu}}%
    \thanks{$^{2}$Department of Electrical Engineering, Sharif University of Technology.}%
}

\begin{document}

\maketitle
\thispagestyle{empty}
\pagestyle{empty}

\input{sec/0_abstract}    
\input{sec/1_intro}
\input{sec/2_related_works}
\input{sec/3_framework}
\input{sec/4_experiments}

\input{sec/5_conclusion}

% References are important to the reader; therefore, each citation must be complete and correct. If at all possible, references should be commonly available publications.

\end{document}

%% file: sec/0_abstract.tex
\begin{abstract}
% Recent vision–language models have evolved to support multi-step reasoning and knowledge integration, particularly for robotics. Despite strong perception abilities, they struggle with higher-order reasoning, particularly spatial relationships. To address these challenges, we propose a novel lightweight multimodal framework that enables accurate, interpretable, and higher-performance understanding of spatial relationships. By integrating visual and symbolic representations, it grounds reasoning in explicit geometric and logical structures, outperforming existing vision–language models.
Visual reasoning, particularly spatial reasoning, is a challenging cognitive task that requires understanding object relationships and their interactions within complex environments, especially in robotics domain. Existing vision–language models (VLMs) excel at perception tasks but struggle with fine-grained spatial reasoning due to their implicit, correlation-driven reasoning and reliance solely on images. We propose a novel neuro-symbolic framework that integrates both panoramic-image and 3D point cloud information, combining neural perception with symbolic reasoning to explicitly model spatial and logical relationships. Our framework consists of a perception module for detecting entities and extracting attributes, and a reasoning module that constructs a structured scene graph to support precise, interpretable queries. Evaluated on the JRDB-Reasoning dataset, our approach demonstrates superior performance and reliability in crowded, human-built environments while maintaining a lightweight design suitable for robotics and embodied AI applications.
\end{abstract}

%% file: sec/1_intro.tex
\section{INTRODUCTION}
Visual reasoning is a challenging cognitive task because it requires not only recognizing objects but also understanding their relationships and interpreting them within complex contexts. Spatial reasoning, in particular, involves inferring the positions, orientations, and interactions of multiple entities, requiring the integration of visual perception with relational and contextual understanding. It is especially demanding for embodied AI systems, such as robots, which must infer both object locations relative to themselves, called Relative Robot Positioning.—for example, determining \textit{which objects are located in front of the robot}. Moreover, in human-built environments, where humans and objects are densely arranged and often interact in complex ways, robots must reason about fine-grained spatial relationships among multiple entities; For instance, identifying \textit{who is positioned at the front left of an elderly person} involves the challenge of determining human orientation and relative spatial positions, a task referred to as relative human positioning.

\begin{figure}[t]
\begin{center}
\scalebox{0.95}{
  \includegraphics[width=1\linewidth]{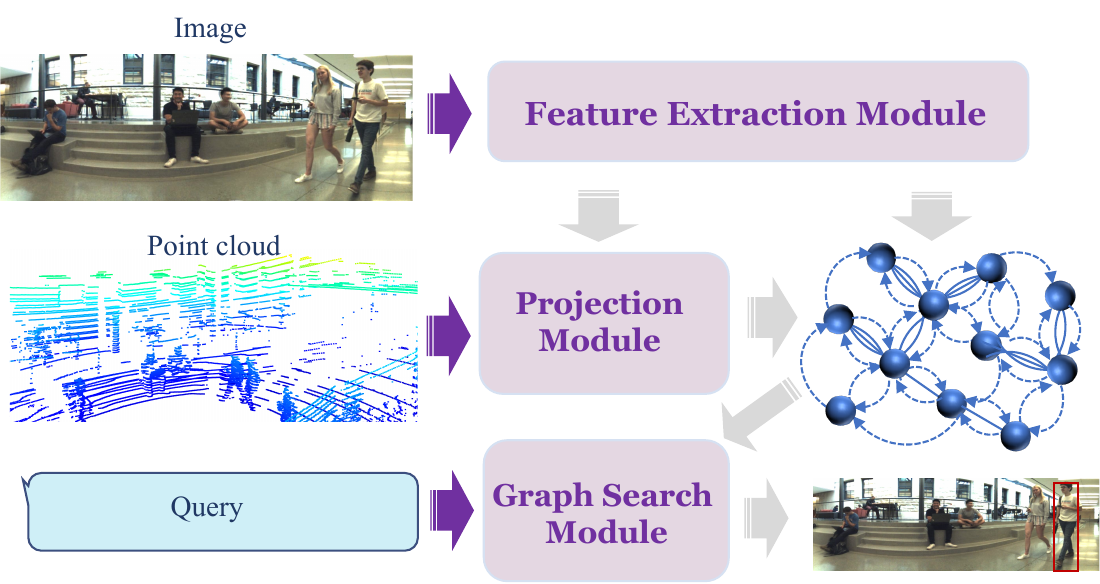}
}
\end{center}
% \vspace{-1em}
\caption{
Schematic of the proposed framework: image, point cloud, and query serve as inputs. Feature extraction and projection generate a graph from attributes and edges, and the graph search module derives the final answer.}
\label{fig:frameworktizer}
% \vspace{-1em}
\end{figure}
% \vspace{-0.5em}

In recent years, vision–language models (VLMs) and compositional models~\cite{suris2023vipergpt,you2023idealgpt,ke2024hydra,cai2025naver} have shown some advancements in visual reasoning problems. However, they have struggled with spatial reasoning, especially in relative human positioning, despite demonstrating strong performance on perception tasks such as object detection~\cite{amizadeh2020neuro, yang2025magicvqa, man2025argus, yi2025corgi, xu2024llavao1, yan2024vigor}. This limitation arises because their reasoning is largely implicit—driven by statistical correlations learned during pretraining—rather than grounded in explicit geometric or logical structures, making them prone to errors in tasks requiring precise relational understanding. Consequently, even state-of-the-art models may hallucinate, produce logically inconsistent answers, or fail to generalize in fine-grained spatial reasoning tasks, limiting their reliability in robotics, navigation, and embodied AI, where accurate spatial understanding is critical. In addition, most current models and frameworks rely solely on non-stitched images for spatial relation problems, neglecting depth and 3D structural information, which further constrains their ability to accurately interpret spatial configurations.

To address these challenges, we proposed a novel neuro-symbolic  lightweight framework that analyzes both stitched-images and 3D information (point clouds), enabling more accurate, interpretable, and higher-performance understanding of spatial relationships for Visual Grounding (VG) and Visual Question Answering (VQA) tasks. Furthermore, by combining visual and symbolic representations, our framework grounds reasoning in explicit geometric and logical structures, reducing errors. This approach overcomes the limitations of state-of-the-art models, providing reliable spatial reasoning critical for robotics, navigation, and other embodied AI applications. 

More specifically, our framework consists of two main components: neural-based perception and symbolic-based reasoning. In the perception part, the Feature Extraction Module leverages foundation vision–language backbones for efficient detection and attribute extraction. Specifically, it first detects people and objects, producing bounding boxes, while InternVL3.5 enriches each detection with attributes such as age, gender, and action. The Projection Module then integrates point cloud data with these extracted features to compute spatial relations between pairs of individuals.

The reasoning component is powered by a novel symbolic graph search module that encodes spatial and logical relationships in a structured form. We construct a scene graph, where nodes represent individuals with their attributes, and edges capture spatial relations derived from bounding boxes, depth cues, and relative positions provided by the projection module. This structured graph acts as an intermediate reasoning layer, enabling precise and interpretable queries, such as “Who is seated to the left of the seated person?” In the graph search module, the query is first converted into a structured format, after which a search algorithm is applied on the graph.

By combining neural perception with symbolic structure, our framework addresses a key limitation of state-of-the-art methods in spatial reasoning.
Moreover, its lightweight design makes it well-suited for robotics and other embodied AI applications. The schematic of our framework is illustrated in Fig.~\ref{fig:frameworktizer}.

We evaluated our proposed framework against state-of-the-art VLMs, and the results demonstrate its strong performance. For this purpose, we employ JRDB-Reasoning~\cite{jahangard2025jrdb}, a robotics dataset built upon JRDB~\cite{martin2021jrdb} and designed for human-built and crowded environments. It provides rich spatial relationships—both relative to the robot and humans—along with multimodal data, including RGB panoramic images and point clouds, and perception-level annotations such as age, race, and gender, enabling advanced 3D perception and reasoning. Its combination of relational and multimodal information offers a level of detail rarely found in existing benchmarks.

In summary, our main contributions are:

\begin{enumerate}
\item Multi-modal lightweight framework: Integrates panoramic images and 3D point clouds for accurate, interpretable spatial reasoning relative to robot and human in crowded environments with few parameters, for robotics.

\item A novel explicit symbolic reasoning: Constructs scene graphs to reduce errors in relative positioning and logical reasoning compared to state-of-the-art VLMs.

\item Achieves superior performance compared to state-of-the-art VLMs.
\end{enumerate}

% \begin{enumerate}
%  \item A neuro-symbolic framework integrating perception and reasoning, combining neural features with symbolic structures for precise and interpretable reasoning.

% \item A multimodal, lightweight framework with fewer parameters than state-of-the-art VLMs, designed for robotic applications.
% \item Achieves superior performance compared to state-of-the-art VLMs.
% \end{enumerate}

%% file: sec/2_related_works.tex
\section{RELATED WORK}
In recent years, Visual Language Models (VLMs) have emerged as a prominent area of research and continue to undergo rapid development~\cite{amizadeh2020neuro,yang2025magicvqa,man2025argus,yi2025corgi,xu2024llavao1,yan2024vigor}. However,
Recent research highlights significant limitations in Vision–Language Models (VLMs) when it comes to spatial reasoning, an ability that is crucial for robotics, embodied AI, and scene understanding. While VLMs have demonstrated impressive performance on general perception and language grounding tasks, they often fail when queries require fine-grained spatial interpretation such as relative positioning, orientation, or multi-object relations. For example, Chen et al.~\cite{chen2024spatialvlm} introduced SpatialVLM, which leverages large-scale synthetic datasets specifically curated to improve metric spatial reasoning. Although this approach improves performance on distance-related tasks, it still underperforms on directional and relational queries such as determining whether an object is “to the left of” or whether an agent is “facing” another entity. In a broader evaluation, Stogiannidis et al.~\cite{stogiannidis2025mindthegap} systematically benchmarked 13 widely used VLMs and reported that their accuracy on spatial reasoning benchmarks was close to random guessing, emphasizing the severity of the problem and the need for fundamentally improved methods.

To address these limitations, several recent approaches have attempted architectural modifications. For instance, Spatial-LLaVA~\cite{sun2025spatialllava} and Spatial-MLLM~\cite{wu2025spatialmllm} introduce improved alignment mechanisms and dual-encoder architectures, which yield measurable gains. However, these models remain grounded primarily in 2D feature representations, and thus lack the ability to reason over explicit 3D geometric structures. Other lines of work attempt to mitigate this by incorporating depth cues or 3D priors: SpatialRGPT~\cite{cheng2024spatialrgpt} and SpatialPIN~\cite{ma2024spatialpin}, for example, extend VLMs with auxiliary modalities to capture richer geometric signals. While promising, these models tend to focus narrowly on pairwise object–object relations and are not designed for reasoning in multi-view or scene-level settings, where viewpoint consistency and relational composition are critical.

Complementing model development, benchmark efforts such as SpatialEval~\cite{wang2024spatialeval} and qualitative reasoning datasets like QSR-Benchmark~\cite{li2024qsrbenchmark} provide further evidence of the gap. These benchmarks consistently demonstrate that current models fail to generalize to viewpoint-dependent reasoning, multi-hop relational queries, and dynamic scene interpretations. Mechanistic interpretability studies also shed light on the underlying causes: Wu et al.~\cite{wu2025interpretability} show that failures in spatial reasoning often arise not from insufficient data exposure but from misaligned attention mechanisms that prioritize surface-level correlations over geometric structure.

Collectively, these works reveal a clear research gap: there is currently no framework that seamlessly integrates explicit geometric modeling, multi-view normalization, and relational reasoning in a unified way. Our proposed approach aims to fill this void. By extracting 3D spatial structures from 2D detections and point clouds, and constructing a scene graph that encodes semantic and geometric relations, we enable accurate and interpretable reasoning in complex, stitched-image environments. This combination of explicit geometry with symbolic reasoning provides a pathway toward high-performance spatial understanding that is both robust and generalizable across diverse real-world tasks.

%% file: sec/3_framework.tex
\section{METHOD}
\begin{figure*}[t]
\begin{center}
\scalebox{0.95}{
  \includegraphics[width=1\linewidth]{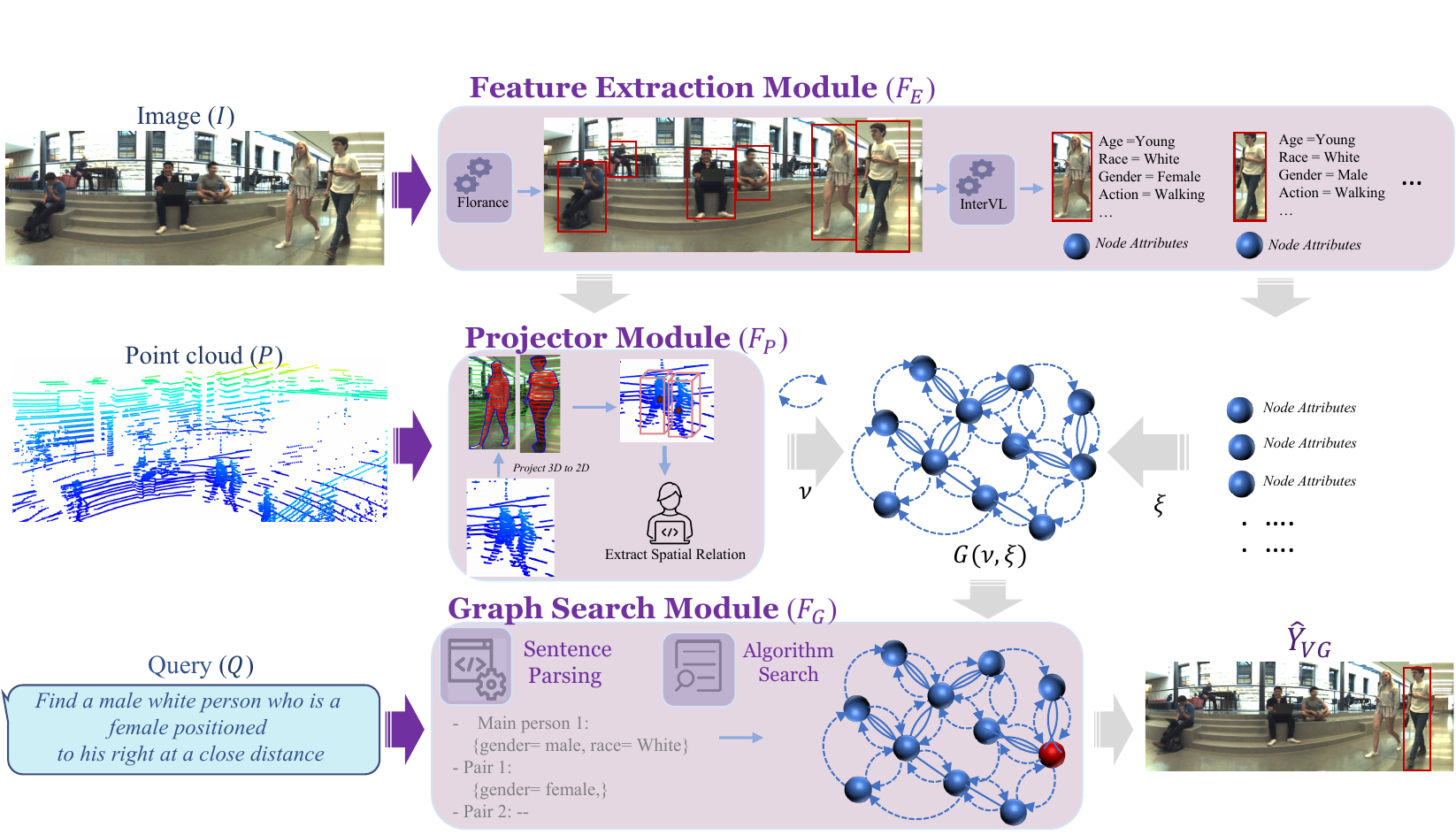}
}
\end{center}
\caption{
The overall framework of the model comprises a perception part—consisting of the \textcolor{mypurple}{Feature Extraction Module ($\mathcal{F}_E$)} and \textcolor{mypurple}{Projection Module ($\mathcal{F}_P$)} for semantic and geometric information extraction—and a reasoning part, the \textcolor{mypurple}{Graph Search Module ($\mathcal{F}_G$)}, which enables query-driven, interpretable spatial understanding.
}
\label{fig:framework}
\end{figure*}
The design of our framework, shown in Figure~\ref{fig:framework}, comprises two main components: a perception and a reasoning part. The perception part includes the \textcolor{mypurple}{\textbf{Feature Extraction Module} ($\mathcal{F}_E$)} and the \textcolor{mypurple}{\textbf{Projection Module} ($\mathcal{F}_P$)}, which work together to extract semantic and geometric information from multimodal inputs. The reasoning part is a \textcolor{mypurple}{\textbf{Graph Search Module} ($\mathcal{F}_G$)} that performs query-driven reasoning over a structured graph, enabling precise and interpretable spatial understanding. 

The framework takes as input a stitched image, a point cloud, and an associated query, denoted as $X = \{I,P,Q\}$, where $I$ is the stitched image, $P$ the point cloud and $Q$ the query. The final output, $\hat{Y}$, may be textual answers for VQA or bounding boxes for the VG task.  

First, the Feature Extraction Module applies pretrained vision-language encoders to the stitched image, detecting and localizing entities,  while extracting rich semantic attributes such as 
age, gender, or actions.
% Florence ~\cite{xiao2023florence} 
% ,InternVL3.5~\cite{wang2025internvl3}). 
Meanwhile, the point cloud provides geometric and spatial structure of the scene. Next, the Projector Module integrates these two streams by aligning semantic features with the geometric relations derived from the point cloud, embedding entities into a structured graph representation. This graph encodes both appearance-based attributes and spatial context. Finally, the Graph Search Module performs query-driven reasoning over this graph: given a query $Q$, the module traverses the graph to identify the most relevant nodes and outputs either bounding boxes or textual descriptions corresponding to the query. 
% First, the Feature Extraction Module applies pretrained vision-language encoders (e.g., Florence~\cite{xiao2023florence} and InternVL3.5~\cite{wang2025internvl3}) to the stitched image, detecting and localizing entities while extracting rich semantic attributes such as age, gender, or actions. Meanwhile, the point cloud provides geometric and spatial structure of the scene. Next, the Projector Module integrates these two streams by aligning semantic features with the geometric relations derived from the point cloud, embedding entities into a structured graph representation. This graph encodes both appearance-based attributes and spatial context. Finally, the Graph Search Module performs query-driven reasoning over this graph: given a query $Q$, the module traverses the graph to identify the most relevant nodes and outputs either bounding boxes or textual descriptions corresponding to the query. 
\subsection{\normalfont Feature Extraction Module}
As part of the perception component, the Feature Extraction Module ($\mathcal{F}_E$) analyzes the stitched image to identify entities of interest. Leveraging pretrained vision–language encoders, it performs object detection and localization via Florence~\cite{xiao2023florence}, producing 2D bounding boxes and segmentation
\(b_i \in \mathbb{R}^{4}\) for each detected entity, and semantic attribute extraction via InternVL3.5~\cite{wang2025internvl3}, producing semantic feature vectors 
\(s_i \in \mathbb{R}^{d_s}\) that encode attributes such as age, gender, and actions, etc.

Mathematically, the module is defined as:
\[
\mathcal{F}_E : I \mapsto \{(b_i, s_i)\}_{i=1}^{M},
\]
where \(I \in \mathbb{R}^{H \times W \times 3}\) is the input stitched image and \(M\) is the number of detected entities. For each entity \(i\), we have:
\[
(b_i, s_i) = \mathcal{F}_E(I)_i.
\]

The resulting semantic attributes serve as descriptors of graph nodes and are later fused with geometric features from the Projection Module to construct the graph. This integration provides both localized visual evidence and high-level semantics, ensuring a discriminative representation of each scene element.

\subsection{\normalfont Projection Module}
The Projection Module ($\mathcal{F}_P$) serves as the bridge between semantic features from the image and geometric information from the point cloud. Given the detected entities \(\{(b_i, s_i)\}_{i=1}^{M}\) from the Feature Extraction Module and a 3D point cloud \(P \in \mathbb{R}^{N \times 3}\), the module establishes geometry-aware correspondences and computes spatial relations between entities.

Using the camera intrinsics \(K \in \mathbb{R}^{3 \times 3}\) and extrinsics \([R|t] \in \mathbb{R}^{3 \times 4}\), each 3D point \(p_j \in P\) is projected onto the image plane:
\[
u_j = K [R|t] \, p_j, \quad u_j \in \mathbb{R}^2.
\]

For each detected entity \(i\), the corresponding 3D points are selected as:
\[
\pi(b_i) = \{p_j \in P \mid u_j \in b_i\}.
\]

The 3D center of the entity is then computed as the mean of its associated points:
\[
c_i = \frac{1}{|\pi(b_i)|} \sum_{p_j \in \pi(b_i)} p_j.
\]

Around this center, a 3D bounding box \(B_i^{3D}\) is defined to capture the volumetric extent of the entity. Pairwise spatial relations between entities \(i\) and \(j\) are encoded as:
\[
r_{ij} = f_\text{geo}(c_i, c_j, B_i^{3D}, B_j^{3D}),
\]
where \(f_\text{geo}\) can include distance, relative orientation, adjacency, and occlusion patterns.

Finally, the unified scene graph is constructed as:
\[
G = (\mathcal{V}, \mathcal{E}), \quad 
\mathcal{V} = \{v_i = (s_i, c_i)\}_{i=1}^{M}, \quad
\mathcal{E} = \{r_{ij}\}_{i,j=1}^{M}.
\]

This graph integrates both semantic and geometric information, providing a structured representation of the scene that is used by the Graph Search Module for query-driven reasoning.

% \subsection{\normalfont Graph Search Module}
% The Graph Search Module ($\mathcal{F}_G$) represents the neuro-symbolic reasoning component of the framework. It performs structured reasoning over the multimodal graph constructed by the Projection Module. Given a textual query $Q$, the query is first analyzed to identify which attributes and relations the graph must consider during the search. The module then formulates the search as a traversal problem over the graph. Nodes represent entities enriched with semantic and geometric attributes, while edges encode relations between entities. The query conditions are matched against node attributes and relational structures, enabling the module to identify the most relevant nodes with respect to the query. For example, a query such as ``young male walking'' leads the module to locate nodes whose attributes satisfy these conditions and whose spatial relations are consistent with the scene. The final output is either a bounding box indicating the localized entity in the image or a textual answer describing the matched result. By combining relational reasoning with attribute filtering, the Graph Search Module provides robust query-driven grounding in complex visual environments.
\subsection{\normalfont Graph Search Module}
The Graph Search Module ($\mathcal{F}_G$) consists of two main submodules: (1) sentence parsing and (2) search algorithm. It represents the reasoning component of the framework, performing query-driven neuro-symbolic reasoning over the graph $G = (\mathcal{V}, \mathcal{E})$ generated by the Projection Module.
\[
\mathcal{F}_G : (G, Q) \mapsto \hat{Y},
\] 
where \(\hat{Y}\) is either a set of bounding boxes for visual grounding VG or a textual answer for VQA tasks. For VG tasks, in this work, the output bounding boxes are:
\[
\hat{Y}_\text{VG} = \{B_i^{2D} \mid v_i \in \hat{\mathcal{V}}\}.
\]
\textbf{Sentence Parsing}.    
Given a textual query \(Q\), this submodule analyzes its structure and extracts semantically meaningful elements relevant for graph-based reasoning. It identifies entities, their attributes, and the relations between them, converting complex natural language queries into structured representations suitable for graph search.  

For example, as shown in Figure~\ref{fig:framework}, the query \emph{``Find a male white person who is a female positioned to his right at a close distance.''} is parsed by identifying the main person with attributes gender = male and race = White, and the related person with attributes gender = female, race = White, and the relational constraints indicating that she is positioned to the right at a close distance.  

This decomposition clearly separates the primary entity, secondary entities, and their relational constraints, enabling precise and efficient reasoning over the graph.
\begin{figure}[t]
\begin{center}
\scalebox{0.95}{
  \includegraphics[width=1\linewidth]{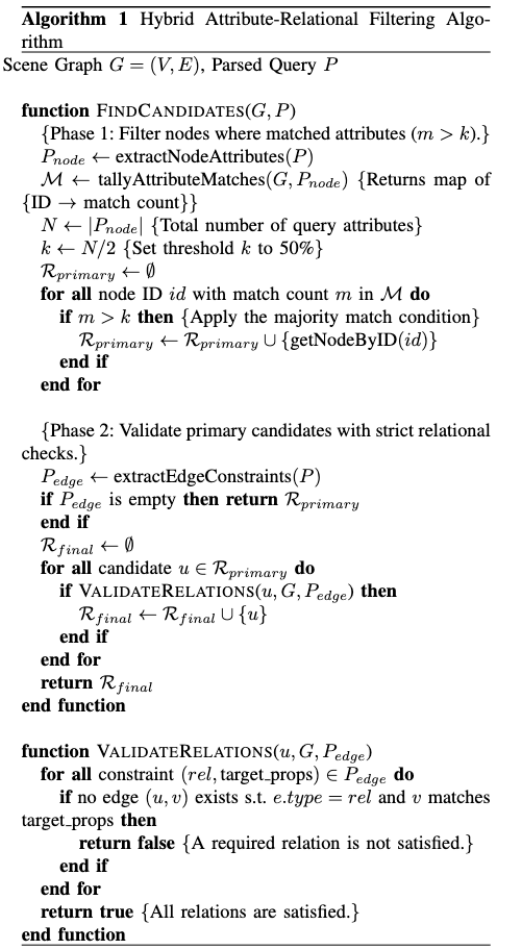}
}
\end{center}
% \vspace{-1em}
\caption{Search Algorithm}
\label{fig:tizer_details}
% \vspace{-1em}
\end{figure}
\noindent\textbf{Search Algorithm}.
We propose a Hybrid Attribute--Relational Filtering Algorithm, which operates in two sequential phases to achieve both robust candidate selection and accurate relational reasoning. The two-phase architecture balances efficiency and accuracy. Phase~1 provides a robust, noise-tolerant filtering mechanism that narrows the candidate set without prematurely discarding valid nodes. Phase~2 enforces strict relational reasoning to resolve complex queries accurately. Together, these phases allow the algorithm to handle both attribute-based and relation-based queries in a unified framework, leveraging the strengths of both approaches while mitigating their individual weaknesses. The algorithm is designed to efficiently process queries that involve both intrinsic node attributes and relational constraints in a scene graph.  

\subsubsection*{Phase 1: Node Attribute Filtering}  
The first phase aims to select a set of primary candidates based solely on the intrinsic attributes of individual nodes, such as age, gender, race, and action. This stage is critical because it reduces the search space early, allowing the system to focus computational resources on plausible candidates rather than the entire set of nodes.  

We employ a \textit{Majority Match} strategy, wherein a node is retained if it satisfies more than half of the attributes specified in the query. This approach provides resilience against minor perception errors that can arise from imperfect detection or classification. For example, a node predicted as ``young adult'' may still be considered a match for a query specifying ``adolescent'' if it satisfies the majority of other attributes.  

Moreover, this phase ensures efficiency by filtering out nodes that fail to meet the majority of attribute requirements. By doing so, it avoids unnecessary relational checks on clearly irrelevant nodes. The threshold of satisfying more than 50\% of the attributes strikes a balance between tolerance to noise and maintaining discriminatory power, ensuring that only plausible candidates advance to the next stage.  

\subsubsection*{Phase 2: Relational Validation}  
The second phase enforces the relational constraints specified in the query. For each candidate identified in Phase~1, we examine its outgoing edges in the scene graph to verify whether it satisfies the required relationships. A candidate is retained only if it is connected to another node through the specified relationship type (e.g., ``to the left of,'' ``interacting with'') and the connected node matches the required target attributes.  

This phase ensures that the final set of candidates is not only individually plausible but also contextually consistent. By explicitly traversing the graph edges, the algorithm enforces structural reasoning that goes beyond simple attribute matching. This is essential for complex queries where the relationship between nodes determines the correct answer, such as spatial relations, interactions, or hierarchical dependencies.  

The relational validation phase effectively compensates for any ambiguity left after the attribute filtering phase. Even if multiple nodes share similar attributes, only those that satisfy the relational constraints are selected, ensuring the accuracy of the final results.

By combining attribute filtering, relational reasoning, and visual-symbolic representations, the Graph Search Module enables precise, interpretable, and query-driven grounding over complex scenes, providing a high-performance understanding of spatial relationships grounded in explicit geometric and logical structures, and effectively overcoming the limitations of VLMs.

%% file: sec/4_experiments.tex
\section{EXPERIMENTS}
Vision-Language Models (VLMs) combine visual perception with language understanding and have shown strong performance in semantic grounding tasks. An important aspect of their capability is reasoning, particularly in understanding spatial relations. In this part, we evaluate VLMs on both perception and spatial reasoning tasks and compare their performance with our proposed framework.
\subsection{\normalfont Dataset and Evaluation Metric}
For our experiments, we use the JRDB-Reasoning benchmark~\cite{jahangard2025jrdb}, which extends the JRDB dataset~\cite{martin2021jrdb} with enriched annotations, an adaptive query engine, and step-by-step reasoning trajectories for evaluating visual reasoning tasks of varying complexity. For visual grounding (VG), we follow the evaluation protocol from prior work~\cite{suris2023vipergpt, ke2024hydra}, where a predicted bounding box is considered correct if its mIoU with the ground truth exceeds 0.5. We also report mAP@[.50:.95], the average mean average precision computed across IoU thresholds from 0.50 to 0.95, providing a comprehensive assessment of detection performance.
\begin{table}[t]
\centering
\begin{tabular}{l c}
\toprule
\textbf{Model} & \textbf{Num. Parameters$\downarrow$} \\
\midrule
LLMDet~\cite{fu2025llmdet}                    & 0.34B      \\
Qwen2.5-VL-Instruction~\cite{bai2025qwen2} & 3B      \\
Qwen2.5-VL-Instruction~\cite{bai2025qwen2} & 7B      \\
MM-G-DINO~\cite{zhao2024open}          & 0.34B \\
Ovis2.5~\cite{lu2025ovis2}                & 2B      \\
Ovis2.5~\cite{lu2025ovis2}                & 9B      \\
\textbf{Our Framework}    & 1.3B      \\
\bottomrule
\end{tabular}
\caption{Comparison of VLM parameters ($\downarrow$ = smaller).}
\label{table:para}
\end{table}
\vspace{-0.5em}

\newcommand{\increase}[2]{\textsuperscript{\textcolor{green}{$\blacktriangle$+#1\%}}#2}
\newcommand{\decrease}[2]{\textsuperscript{\textcolor{red}{$\blacktriangledown$-#1\%}}#2}

\begin{table*}[t!]
\caption{Performance comparison on various attribute detection tasks. We report mAP@[.5,.95] and mIOU metrics. Superscripts show the percentage change of our model compared to the best-performing baseline in each category.}
\label{tab:performance_comparison_baseline}
\centering

% --- TABLE (b): Second set of attributes ---
\begin{subtable}{1.0\textwidth}
    \centering
    \caption{Results for attributes including distance and spatial relations like distance-SR/robot,HHG.}
    \label{tab:part2}
    \begin{tabular}{@{}l rr rr rr@{}}
    \toprule
    & \multicolumn{2}{c}{\textbf{gender-age-race-distance}}
    & \multicolumn{2}{c}{\textbf{gender-age-race-distance-SR/robot}}
    & \multicolumn{2}{c}{\textbf{gender-age-race-HHG}} \\
    \cmidrule(r){2-3} \cmidrule(lr){4-5} \cmidrule(l){6-7}
    & mAP & mIOU
    & mAP & mIOU
    & mAP & mIOU \\
    \midrule
    Ovis2.5-2B~\cite{lu2025ovis2}   & 2.83  & 6.23  & 2.83  & 5.97  & 4.90  & 13.80 \\
    Ovis2.5-9B~\cite{lu2025ovis2}   &  \textbf{12.18}   &   \textbf{22.42}    &   \textbf{8.2}    &   \textbf{14.1}    &    12.60   &    26.34  \\
    LLMDet~\cite{fu2025llmdet}    & 12.02 & 21.50 & 5.96 & 11.97 & 11.20 & 24.30 \\
    MM-G-DINO~\cite{zhao2024open}    & 8.08  & 18.32 & 3.33  & 8.24  & \textbf{14.40} & \textbf{28.20} \\
    Qwen2.5-VL-3B~\cite{bai2025qwen2}   & 0.60  & 5.68  & 0.20  & 3.38  & 1.55  & 11.62 \\
    Qwen2.5-VL-7B~\cite{bai2025qwen2}   & 6.92  & 12.90  & 2.51  & 6.54  &  13.16     &   20.03    \\
    \midrule
    \textbf{Our Framework}      & \increase{35.7}\textbf{16.53} & \increase{18.2}\textbf{26.50} & \increase{96.7}\textbf{16.13} & \increase{67.8}\textbf{23.66} &   \increase{43.0}\textbf{20.60}  &   \increase{35.1}\textbf{38.09} \\
    \bottomrule
    \end{tabular}
\end{subtable}
\vspace{0.5cm} 

% --- TABLE (a): First set of attributes ---
\begin{subtable}{1.0\textwidth}
    \centering
    \caption{Results for [human], [age], [gender-age], and [gender-age-race] attributes.}
    \label{tab:part1}
    \begin{tabular}{@{}l rr rr rr rr@{}}
    \toprule
    & \multicolumn{2}{c}{\textbf{human}}
    & \multicolumn{2}{c}{\textbf{age}}
    & \multicolumn{2}{c}{\textbf{gender-age}}
    & \multicolumn{2}{c}{\textbf{gender-age-race}} \\
    \cmidrule(r){2-3} \cmidrule(lr){4-5} \cmidrule(lr){6-7} \cmidrule(l){8-9}
    & mAP & mIOU & mAP & mIOU & mAP & mIOU & mAP & mIOU \\
    \midrule
    Ovis2.5-2B~\cite{lu2025ovis2}   & 4.79  & 19.11 & 6.63  & 28.69 & 6.13  & 24.75 & 4.18  & 16.5  \\
    Ovis2.5-9B~\cite{lu2025ovis2}   &   23.24    &   48.35    &   24.61    &    43.20   &  \textbf{20.36}     &    34.2   &   \textbf{14.32}   &   \textbf{24.38}    \\

    LLMDet~\cite{fu2025llmdet}   & \textbf{47.36} & \textbf{67.65} & \textbf{34.97} & 52.84 & 16.90 & 37.36 & 13.37 & 23.0 \\
    MM-G-DINO~\cite{zhao2024open}   & 24.81 & 55.16 & 28.87 & \textbf{53.10} & 19.48 & \textbf{41.46} & 10.09 & 22.6  \\
    Qwen2.5-VL-3B~\cite{bai2025qwen2}   & 8.07  & 29.69 & 7.29  & 27.53 & 3.88  & 19.8  & 2.40  & 8.78  \\
    Qwen2.5-VL-7B~\cite{bai2025qwen2}   & 16.65 & 35.41 & 18.39 & 35.74 & 15.95 & 28.43 & 10.21 & 17.1  \\
    \midrule
    \textbf{Our Framework}       & \increase{15.1}\textbf{54.49} & \increase{6.9}\textbf{72.31} & \increase{15.0}\textbf{40.20} & \increase{9.6}\textbf{58.21} & \increase{9.3}\textbf{22.26} & \decrease{13.6}{35.81} & \decrease{14.4}{12.26} & \decrease{18.4}{19.9} \\
    \bottomrule
    \end{tabular}
\end{subtable}

\label{table:r1}
\end{table*}
% , which we categorize into three geometric difficulty levels, denoted by $\mathcal{G}_1$, $\mathcal{G}_2$, and $\mathcal{G}_3$, defined as
% {\small
% $\mathcal{G}_1 = { (S, R, T) \mid S = 1, R = 0},;
% \mathcal{G}_2 = { (S, R, T) \mid S = 2, R \geq 1},;
% \mathcal{G}_3 = { (S, R, T) \mid S = 3, R \geq 2},$
% }
% where $S$ denotes the number of nodes (geometric entities), $R$ represents the number of interactions (edges), and $T$ indicates the number of time slots that we consider it one (image).\\

\begin{table*}[t!]
\caption{Performance comparison on various attribute detection tasks. We report mAP@[.5,.95] and mIOU metrics. Superscripts show the percentage change of each model compared to the BASE model.}
\label{tab:performance_comparison}
\centering
\begin{tabular}{@{}l rr rr rr rr@{}}
\toprule
& \multicolumn{2}{c}{\textbf{gender-age-race}}
& \multicolumn{2}{c}{\textbf{gender-age-race-distance}}
& \multicolumn{2}{c}{\textbf{gender-age-race-distance-SR/robot}}
& \multicolumn{2}{c}{\textbf{gender-age-race-HHG}} \\
\cmidrule(r){2-3} \cmidrule(lr){4-5} \cmidrule(lr){6-7} \cmidrule(l){8-9}
& mAP & mIOU & mAP & mIOU & mAP & mIOU & mAP & mIOU \\
\midrule
Qwen2.5-VL-3B   & \increase{33.0}{16.3} & \increase{34.4}{26.74} & \increase{10.3}{18.23} & \increase{18.5}{31.4} & \increase{10.2}{17.78} & \increase{12.3}{26.57} & \increase{2.2}{21.06} & \increase{1.8}{38.79} \\
InternVL3\_5-2B & \increase{19.1}{14.6} & \increase{17.4}{23.36} & \increase{6.7}{17.64} & \increase{11.1}{29.43} & \increase{15.7}{18.67} & \increase{13.8}{26.93} & \increase{3.3}{21.28} & \increase{4.3}{39.74} \\
InternVL3\_5-4B & \increase{55.7}{19.09} & \increase{57.8}{31.4} & \increase{35.0}{22.32} & \increase{40.9}{37.33} & \increase{24.9}{20.14} & \increase{32.7}{31.4} & \increase{19.6}{24.63} & \increase{9.8}{41.83} \\
\bottomrule
\end{tabular}
\label{table:r2}
\end{table*}

\begin{figure*}[t]
\begin{center}
\scalebox{0.95}{
  \includegraphics[width=1\linewidth]{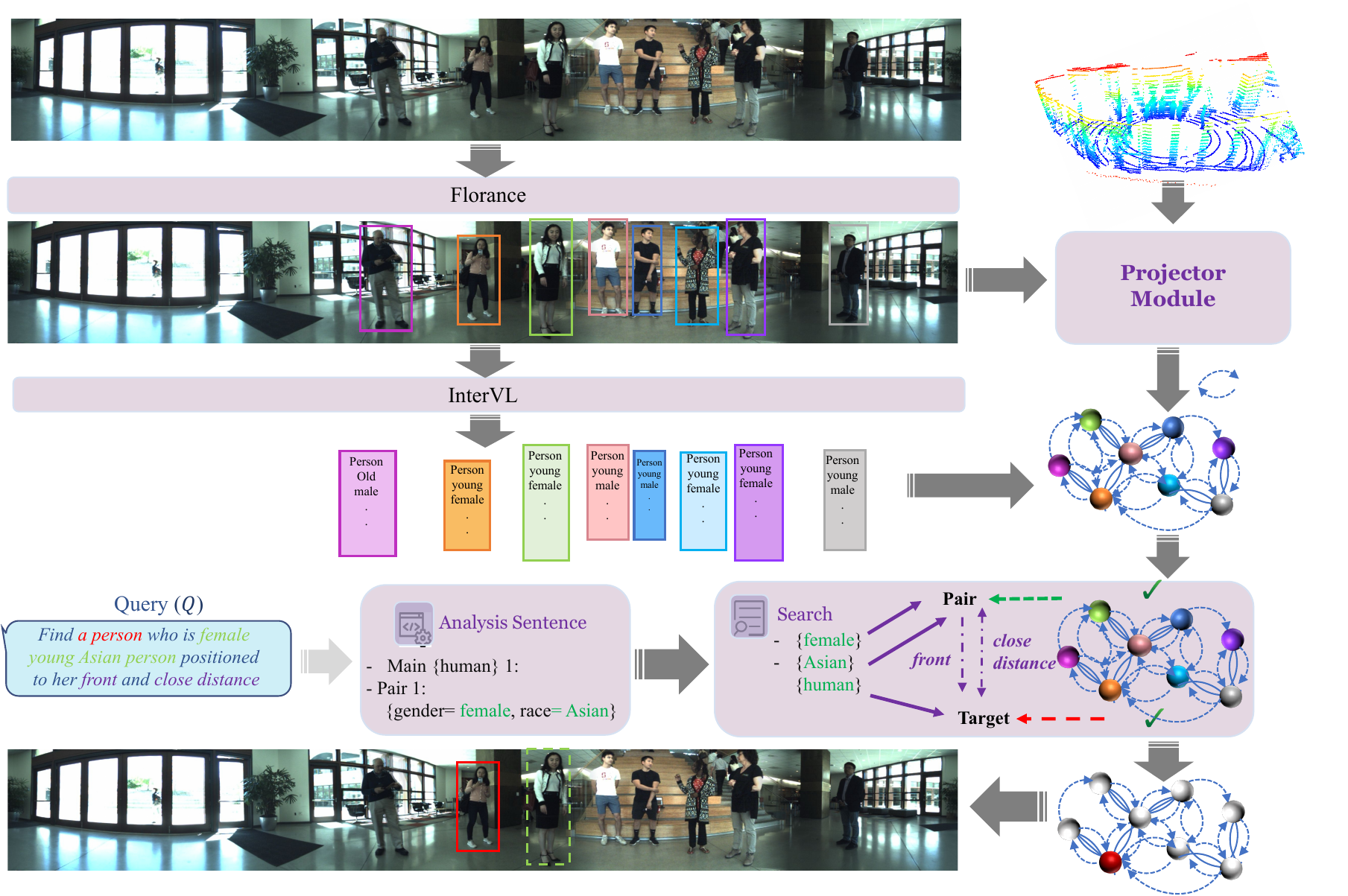}
}
\end{center}
\caption{
A sample of our framework’s pipeline: the stitched image is first processed by Florance to detect humans, and their bounding boxes are passed to InterVL for attribute extraction and graph node construction. Outputs from Florance and the point cloud are integrated to derive spatial relations, and the resulting graph is searched based on the provided symbols.
}
\label{fig:visulize}
\end{figure*}

\subsection{\normalfont Results}
\noindent\textbf{Parameters:}
Table~\ref{table:para} presents a comparative analysis of recent VLMs, focusing on parameter size and reasoning capability. A clear distinction emerges between models designed for specialized perceptual tasks and those engineered for broader cognitive functions. Models such as LLMDet~\cite{fu2025llmdet}, the Qwen2.5-VL variants (3B and 7B)\cite{bai2025qwen2}, and MM-G-DINO~\cite{zhao2024open} operate across a wide parameter range but lack explicit high-level reasoning abilities. In contrast, models specifically built for reasoning, including Ovis2.5 (2B and 9B)\cite{lu2025ovis2} and our proposed framework, demonstrate a more effective integration of scale and reasoning capability.
Notably, our framework delivers robust reasoning with only 1.3B parameters. This positions it as exceptionally efficient, being significantly smaller than the large reasoning-capable Ovis2.5 variant (9B)~\cite{lu2025ovis2}. Furthermore, it introduces this critical capability at a parameter scale where other, even larger models like the Qwen2.5-VL-7B-Instruct, do not. This contrast highlights the efficiency of our proposed design, which enables advanced cognitive functions at a budget closer to lightweight models rather than requiring the scale of massive VLMs.
In summary, the results show that reasoning ability is not solely dependent on parameter count; careful architectural design can achieve reasoning capabilities with smaller models, marking a significant step toward more accessible and efficient VLMs.\\
\textbf{Performance:}
Table~\ref{table:r1} presents a comprehensive comparison of various models on attribute detection and spatial reasoning tasks, evaluated using mAP@[.5,.95] and mIOU metrics.
We evaluated the models at two distinct levels: node level and combined node–edge level.

At the combined node–edge level (Table~\ref{table:r1}(a)), questions involve both node attributes and human–human spatial relations (edges). These queries combine attributes like gender–age–race with distance or orientation relative to the robot, e.g., \textit{``Find the young White male positioned at a close distance relative to me (the robot).''} In another case, the combination of [gender–age–race–HHG] incorporates Human–Human Geometry (HHG), where the focus is on spatial relations between people. For instance: \textit{``Find a White male who has a female positioned to his right at a close distance,''} as illustrated in Figure~\ref{fig:framework}.

At the node level, questions focus on node attributes such as age, gender, race, and their combinations, as shown in Table~\ref{table:r1}(b). For example: \textit{``Find the young people''} or \textit{``Find the females''}.

While baseline models such as LLMDet and MM-G-DINO achieve strong results in selected categories—LLMDet particularly in basic human and age detection, and MM-G-DINO in some mIOU cases—our framework consistently outperforms them in both accuracy and robustness across diverse attribute settings.

In the basic attribute categories, including human, age, [gender-age], and [gender-age-race,] our framework achieves the highest scores in human and age detection, surpassing LLMDet by a significant margin with a 15.1\% increase in mAP and a 6.9\% increase in mIOU for human attributes, and a 15.0\% mAP improvement for age. Even in the challenging [gender-age] and [gender-age-race] settings, where most baselines struggle, our framework maintains competitive performance, demonstrating its strong capacity to generalize across multi-attribute detection.

More importantly, in the complex relational categories involving distance, spatial relations, and human-human/robot interactions, our framework shows clear dominance over all baselines. It achieves the largest relative improvements, such as a remarkable 96.7\% gain in mAP and 67.8\% gain in mIOU for [gender-age-race-distance-SR/robot], and strong advances in human-human interaction detection with 43.0\% higher mAP and 35.1\% higher mIOU compared to the best-performing baselines. These results underscore our model’s unique strength in capturing relational and spatial reasoning, where prior models are limited.

Overall, while specialized models like LLMDet and MM-G-DINO occasionally lead in isolated metrics, our framework delivers the most balanced and superior performance across both basic and complex attributes, setting a new state-of-the-art in attribute detection and relational reasoning.

A key advantage of our framework is its modular, task-agnostic design, which allows for the flexible integration of various model backbones. To quantify this performance enhancement, we conducted a targeted experiment focusing on complex reasoning categories, including fine-grained attributes (gender, age, race) and challenging spatial relations (e.g., distance-SR/robot, HHG). For this evaluation, we identified Qwen2.5-VL-3B-Instruct as a model with suboptimal baseline performance, as shown in Table~\ref{table:r1}. By integrating this specific model as the primary backbone within our system, we demonstrate the framework's ability to significantly uplift its capabilities in these demanding areas. The results of this enhancement are presented in Table~\ref{table:r2}, showcasing the substantial value our architecture adds.

%% file: sec/5_conclusion.tex
\section{CONCLUSIONS}
In this work, we presented a lightweight multimodal  neuro-symbolic framework that explicitly separates perception from reasoning to overcome the limitations of current vision–language models in fine-grained spatial reasoning. By integrating neural perception with a neuro-symbolic reasoning module, our approach grounds inference in explicit geometric and logical structures. Unlike large-scale VLMs that rely primarily on implicit statistical correlations, our framework offers interpretable, structured, and reliable reasoning while remaining computationally efficient. This makes it particularly suitable for embodied AI applications. Furthermore, our results show that the framework not only achieves superior performance compared to state-of-the-art models but also requires significantly fewer parameters.\\
\noindent\textbf{Limitation.} The attribute extraction is constrained by the current capabilities of the LLMs. Improving these models could be a promising direction for future work.

%% file: root.bbl
\begin{thebibliography}{99}

\bibitem{c1} A. Radford, J. W. Kim, C. Hallacy, et al., “Learning transferable visual models from natural language supervision,” in *Proc. Int. Conf. Mach. Learn. (ICML)*, 2021, pp. 8748–8763.

\bibitem{c2} J. Li, D. Li, S. Savarese, and S. C. Hoi, “BLIP: Bootstrapping Language-Image Pre-training for Unified Vision-Language Understanding and Generation,” in *Proc. Int. Conf. Mach. Learn. (ICML)*, 2022, pp. 12888–12900.

\bibitem{c3} X. Chen, et al., “InternVL: Scaling up Vision Foundation Models and Aligning for Generic Vision-Language Representation,” in *Proc. IEEE/CVF Conf. Comput. Vis. Pattern Recognit. (CVPR)*, 2024.  

\bibitem{c4} Y. Chen, et al., “Florence-VL: Enhancing Vision-Language Models with Generative Vision Encoder and Depth-Breadth Joint Training,” in *Proc. IEEE/CVF Conf. Comput. Vis. Pattern Recognit. (CVPR)*, 2025. 

\bibitem{c5} Y. Liu, et al., “SpatialVLM: Endowing Vision-Language Models with Spatial Reasoning Capability,” *arXiv preprint* arXiv:2401.12168, 2024. 

\bibitem{c6} Z. Li, et al., “InternSpatial: Scaling Up Spatial Reasoning in Vision-Language Models,” *arXiv preprint* arXiv:2506.18385, 2025. 

\bibitem{c7} H. Zhou, et al., “Beyond Semantics: Rediscovering Spatial Awareness in Vision-Language Models,” *arXiv preprint* arXiv:2503.17349, 2025. 

\bibitem{c8} A. Singh, et al., “Is a Picture Worth a Thousand Words? Delving Into Spatial Reasoning in Vision-Language Models,” *arXiv preprint* arXiv:2406.14852, 2024.

\bibitem{internvl2024} X. Chen, et al., “InternVL: Scaling up Vision Foundation Models and Aligning for Generic Vision-Language Representation,” in *Proc. IEEE/CVF Conf. Comput. Vis. Pattern Recognit. (CVPR)*, 2024.  

\bibitem{florencevl2025} Y. Chen, et al., “Florence-VL: Enhancing Vision-Language Models with Generative Vision Encoder and Depth-Breadth Joint Training,” in *Proc. IEEE/CVF Conf. Comput. Vis. Pattern Recognit. (CVPR)*, 2025.  

\bibitem{spatialvlm2024} Y. Liu, et al., “SpatialVLM: Endowing Vision-Language Models with Spatial Reasoning Capability,” *arXiv preprint* arXiv:2401.12168, 2024.  

\bibitem{internspatial2025} Z. Li, et al., “InternSpatial: Scaling Up Spatial Reasoning in Vision-Language Models,” *arXiv preprint* arXiv:2506.18385, 2025.  

\bibitem{beyondsemantics2025} H. Zhou, et al., “Beyond Semantics: Rediscovering Spatial Awareness in Vision-Language Models,” *arXiv preprint* arXiv:2503.17349, 2025.  

\bibitem{spatialeval2024} A. Singh, et al., “Is a Picture Worth a Thousand Words? Delving Into Spatial Reasoning in Vision-Language Models,” *arXiv preprint* arXiv:2406.14852, 2024.  

\bibitem{suris2023vipergpt} D. Sur’is, S. Menon, and C. Vondrick, “ViperGPT: Visual Inference via Python Execution for Reasoning,” in *Proc. IEEE/CVF Int. Conf. Comput. Vis. (ICCV)*, 2023, pp. 11854–11864.  

\bibitem{ke2024hydra} F. Ke, Z. Cai, S. Jahangard, W. Wang, P. D. Haghighi, and H. Rezatofighi, “HYDRA: A Hyper Agent for Dynamic Compositional Visual Reasoning,” in *Proc. Eur. Conf. Comput. Vis. (ECCV)*, 2024, pp. 132–149.  

\bibitem{chen2024spatialvlm} C. Chen, et al., “SpatialVLM: Endowing Vision-Language Models with Spatial Reasoning Capabilities,” in \textit{Proc. IEEE/CVF Conf. Computer Vision and Pattern Recognition (CVPR)}, 2024.

\bibitem{bai2025univg} S. Bai, M. Li, Y. Liu, J. Tang, H. Zhang, L. Sun, X. Chu, and Y. Tang, “UniVG-R1: Reasoning guided universal visual grounding with reinforcement learning,” \textit{arXiv preprint arXiv:2505.14231}, 2025.

\bibitem{bai2025qwen2} S. Bai, K. Chen, X. Liu, J. Wang, W. Ge, S. Song, K. Dang, P. Wang, S. Wang, J. Tang, et al., “Qwen2.5-VL Technical Report,” \textit{arXiv preprint arXiv:2502.13923}, 2025.

\bibitem{shakarian2023neuro} 
P. Shakarian, C. Baral, G. I. Simari, B. Xi, L. Pokala, \textit{Neuro Symbolic Reasoning and Learning}, Springer, 2023.

\bibitem{wang2025internvl3} W. Wang, Z. Gao, L. Gu, H. Pu, L. Cui, X. Wei, Z. Liu, L. Jing, S. Ye, J. Shao, et al., “InternVL3.5: Advancing open-source multimodal models in versatility, reasoning, and efficiency,” \textit{arXiv preprint arXiv:2508.18265}, 2025.

\bibitem{fu2025llmdet} S. Fu, Q. Yang, Q. Mo, J. Yan, X. Wei, J. Meng, X. Xie, and W.-S. Zheng, “LLMDet: Learning strong open-vocabulary object detectors under the supervision of large language models,” in \textit{Proc. IEEE/CVF Conf. Computer Vision and Pattern Recognition (CVPR)}, 2025, pp. 14987–14997.

\bibitem{lu2025ovis2} S. Lu, Y. Li, Y. Xia, Y. Hu, S. Zhao, Y. Ma, Z. Wei, Y. Li, L. Duan, J. Zhao, et al., “OVIS2.5 Technical Report,” \textit{arXiv preprint arXiv:2508.11737}, 2025.

\bibitem{stogiannidis2025mindthegap} I. Stogiannidis, et al., “Mind the Gap: Benchmarking Spatial Reasoning in Vision-Language Models,” \textit{arXiv preprint} arXiv:2503.19707, 2025.


\bibitem{you2023idealgpt} H. You, R. Sun, Z. Wang, L. Chen, G. Wang, H. A. Ayyubi, K.-W. Chang, and S.-F. Chang, “IdealGPT: Iteratively Decomposing Vision and Language Reasoning via Large Language Models,” \textit{arXiv preprint arXiv:2305.14985}, 2023.

\bibitem{cai2025naver} Z. Cai, F. Ke, S. Jahangard, M. G. de la Banda, R. Haffari, P. J. Stuckey, and H. Rezatofighi, “Naver: A neuro-symbolic compositional automaton for visual grounding with explicit logic reasoning,” \textit{arXiv preprint arXiv:2502.00372}, 2025.


\bibitem{amizadeh2020neuro} S. Amizadeh, H. Palangi, A. Polozov, Y. Huang, and K. Koishida, “Neuro-symbolic visual reasoning: Disentangling,” in \textit{Proceedings of the International Conference on Machine Learning (ICML)}, pp. 279--290, 2020.

\bibitem{sun2025spatialllava} Y. Sun, et al., “Spatial-LLaVA: Enhancing Large Language Models with Spatial Referring Expressions,” \textit{arXiv preprint} arXiv:2505.12194, 2025.

\bibitem{yang2025magicvqa} S. Yang, S. C. Han, S. Luo, and E. Hovy, “MAGIC-VQA: Multimodal And Grounded Inference with Commonsense Knowledge for Visual Question Answering,” \textit{arXiv preprint} arXiv:2503.18491, 2025.

\bibitem{man2025argus} Y. Man, D.-A. Huang, G. Liu, S. Sheng, S. Liu, L.-Y. Gui, J. Kautz, Y.-X. Wang, and Z. Yu, “Argus: Vision-Centric Reasoning with Grounded Chain-of-Thought,” in \textit{Proceedings of the IEEE/CVF Conference on Computer Vision and Pattern Recognition (CVPR)}, pp. 1234--1243, 2025.
\bibitem{wu2025reinforcing} J. Wu, J. Guan, K. Feng, Q. Liu, S. Wu, L. Wang, W. Wu, and T. Tan, “Reinforcing Spatial Reasoning in Vision-Language Models with Interwoven Thinking and Visual Drawing,” \textit{arXiv preprint} arXiv:2506.09965, 2025.

\bibitem{stogiannidis2025mind} I. Stogiannidis, S. A. Tsaftaris, et al., “Mind the Gap: Benchmarking Spatial Reasoning in Vision-Language Models,” \textit{arXiv preprint} arXiv:2503.19707, 2025.

\bibitem{mayer2025ivispar} J. Mayer, M. Ballout, S. Jassim, F. N. Nezami, and E. Bruni, “iVISPAR: An Interactive Visual-Spatial Reasoning Benchmark for VLMs,” \textit{arXiv preprint} arXiv:2502.03214, 2025.

\bibitem{khemlani2025vision} S. Khemlani, et al., “Vision Language Models Are Unreliable at Trivial Spatial Cognition,” \textit{arXiv preprint} arXiv:2504.16061, 2025.


\bibitem{yi2025corgi} S. Yi and L. Shang, “CoRGI: Verified Chain-of-Thought Reasoning with Visual Grounding,” \textit{arXiv preprint} arXiv:2508.00378, 2025.

\bibitem{xu2024llavao1} 
X. Xu, Z. Xie, H. Wu, X. Bai, and J. Xie, 
``LLaVA-o1: Let Vision Language Models Reason Step-by-Step,'' 
\textit{arXiv preprint} \texttt{arXiv:2411.10440}, 2024.

\bibitem{yan2024vigor} S. Yan, M. Bai, W. Chen, X. Zhou, Q. Huang, and L. E. Li, “ViGoR: Improving Visual Grounding of Large Vision Language Models with Fine-Grained Reward Modeling,” in \textit{Proceedings of the European Conference on Computer Vision (ECCV)}, pp. 4567--4576, 2024.


\bibitem{xiao2023florence} B. Xiao, H. Wu, W. Xu, X. Dai, H. Hu, Y. Lu, M. Zeng, C. Liu, and L. Yuan, “Florence-2: Advancing a unified representation for a variety of vision tasks,” \textit{arXiv preprint arXiv:2311.06242}, 2023.

\bibitem{wu2025spatialmllm} J. Wu, et al., “Spatial-MLLM: Boosting MLLM Capabilities in Visual-based Spatial Intelligence,” \textit{arXiv preprint} arXiv:2503.07557, 2025.

\bibitem{cheng2024spatialrgpt} R. Cheng, et al., “SpatialRGPT: Grounded Spatial Reasoning in Vision Language Models,” \textit{arXiv preprint} arXiv:2406.01584, 2024.

\bibitem{ma2024spatialpin} T. Ma, et al., “SpatialPIN: Enhancing Spatial Reasoning Capabilities through Prompting and Interacting 3D Priors,” \textit{arXiv preprint} arXiv:2403.13438, 2024.

\bibitem{wang2024spatialeval} A. Singh, et al., “Is a Picture Worth a Thousand Words? Delving Into Spatial Reasoning in Vision-Language Models,” \textit{arXiv preprint} arXiv:2406.14852, 2024.

\bibitem{zhao2024open} X. Zhao, Y. Chen, S. Xu, X. Li, X. Wang, Y. Li, and H. Huang, “An open and comprehensive pipeline for unified object grounding and detection,” \textit{arXiv preprint arXiv:2401.02361}, 2024.

\bibitem{li2024qsrbenchmark} H. Li, et al., “Reframing Spatial Reasoning Evaluation in Language Models: A Real-World Simulation Benchmark,” \textit{arXiv preprint} arXiv:2405.15064, 2024.

\bibitem{wu2025interpretability} Y. Wu, et al., “Why Is Spatial Reasoning Hard for VLMs? A Mechanistic Interpretability Study,” \textit{arXiv preprint} arXiv:2503.01773, 2025.

\bibitem{jahangard2025jrdb} S. Jahangard, et al., “JRDB-Reasoning: A Difficulty-Graded Benchmark for Visual Reasoning in Robotics,” \textit{arXiv preprint} arXiv:2508.10287, 2025.

\bibitem{martin2021jrdb} R. Martin-Martin, et al., “Jrdb: A dataset and benchmark of egocentric robot visual perception of humans in built environments,” \textit{IEEE Transactions on Pattern Analysis and Machine Intelligence}, vol. 45, no. 6, pp. 6748--6765, 2021.


\end{thebibliography}
